\documentclass{article} % For LaTeX2e
\usepackage{times}
\usepackage[preprint]{neurips_2026}

\usepackage{amsmath,amsfonts,bm}

\def\eqref#1{equation~\ref{#1}}
\def\1{\bm{1}}

\DeclareMathAlphabet{\mathsfit}{\encodingdefault}{\sfdefault}{m}{sl}
\SetMathAlphabet{\mathsfit}{bold}{\encodingdefault}{\sfdefault}{bx}{n}

\usepackage{hyperref}
\usepackage{url}
\usepackage{booktabs}
\usepackage{amsmath,amssymb,graphicx,booktabs}
\usepackage[utf8]{inputenc}
\usepackage[table]{xcolor}
\usepackage{enumitem}
\usepackage{fancyhdr}
\usepackage{comment}
\usepackage{tocloft}

\setlist[itemize]{leftmargin=1.5em, itemsep=0pt, topsep=2pt, parsep=0pt}
\setlist[enumerate]{leftmargin=1.5em, itemsep=0pt, topsep=2pt, parsep=0pt}

\title{From Pixel Generation to Topological Inference: Structural Dual Super-Resolution for Trustworthy Cross-Physical-Domain Trabecular Morphology Learning}

\author{Fan Zhang, Yi Zhang, Yuhui Dong \\%\&*\\  \thanks{ Use footnote for providing further information about author (webpage, alternative address)---\emph{not} for acknowledging funding agencies.  Funding acknowledgements go at the end of the paper.} \\
Institute of High Energy Physics, Chinese Academy of Sciences \\
Beijing 100049, China\\
\texttt{\{zhangfan95,zhangyi88, dongyh\}@ihep.ac.cn} \\
\And
Ling Wang\\
Department of Radiology, Beijing Jishuitan Hospital\\Beijing, China\\
\texttt{\{1988yisheng\}@163.com} \\
\And
Xin Xu\\
Weifang Community Health Service Center, Pudong New Area\\ Shanghai, China\\
\texttt{\{xuxin211\}@126.com} \\
}
\newlistof{appendix}{app}{Appendix}

\newcommand{\appsection}[1]{%
  \section{#1}%
  \addcontentsline{app}{section}{\protect\numberline{\thesection}#1}%
}

\begin{document}

\maketitle

%\begin{comment}
\vspace{-1em}
\begin{center}
\small
\textcopyright\ 2026 The Authors. All rights reserved. No part of this preprint may be reproduced, distributed, or reused in any form without the prior written permission of the authors.
\end{center}
\vspace{1em}
%\end{comment}

\begin{abstract}

\textbf{Purpose:} To achieve trustworthy macro-to-micro structural inference on UHRCT, reaching the gold-standard 3D bone microstructure morphological assessment of synchrotron radiation micro-CT.

\textbf{Materials and Methods:} UHRCT and SR$\mu$CT are fundamentally different physical imaging domains, resulting in a 31.25-fold difference in resolution; they can only be roughly matched due to differences in data volume, as well as partial volume effects, noise, and beam hardening/scattering artifacts specific to UHRCT. Because existing pixel-generation super-resolution methods do not explicitly model these physical differences, they underperform in this setting. We therefore reframed the task as topological inference rather than pixel generation: deterministic prediction of invariant microstructures from macro-scale inputs, evaluated by morphological parameters. To implement this, we designed a structural dual network that coupled forward physical degradation (micro-to-macro) with inverse structural inference (macro-to-micro), together with pyramid structural consistency discriminators and four structural duality constraints. This network achieved morphology learning as cross-physical-domain consistent structure mining and thereby ensured reconstruction reliability.

\textbf{Results:} SDN achieved morphological parameters largely consistent with SR$\mu$CT across seven metrics. Clinical UHRCT alone showed large errors (BV/TV, 0.3793; 228\% error; Tb.Th, 357.2 $\mu$m; 4793\% error). SDN enabled UHRCT micro-imaging-level quantification (BV/TV, 0.1156 vs 0.1163; Tb.Th, 7.3 $\mu$m vs 10 $\mu$m), with SSIM 0.8. Trained on 3.2 $\mu$m SSRF data, SDN generalized to 3.25 $\mu$m isotropic BSRF data from an independent source.

\textbf{Conclusion:} Topology-inference super-resolution enabled clinical UHRCT to achieve mesoscale-level morphological quantification and cross-source generalization, establishing trustworthy structural inference rather than pixel generation. This paradigm shift demonstrates that topology inference can address the extreme ill-posed problem caused by the ultrahigh resolution gap, few-shot supervision, and the difficulty of strict pairing, offering a new perspective for the research community.
\end{abstract}

\section{Introduction}
\label{sec:intro}

Osteoporosis, a state of systemically reduced bone mineralization and degraded microarchitecture and hence increased risk of fractures, is a global public health issue, with rising prevalence~\cite{ziegeler2026ai}. Bone mineral density (BMD), as measured by dual-energy X-ray absorptiometry (DXA), remains one of the most important indicators for predicting the risk of fragility fractures \cite{10.1359/JBMR.050304}; it is also the diagnostic criterion for osteoporosis in most international guidelines and the therapeutic target for pharmacological interventions \cite{Haseltine446}. Besides, bone microstructure assessment is a key indicator of osteoporosis. The WHO/IOF definition lists “low bone mass” and “bone microstructural deterioration” as core characteristics, indicating that bone microstructure assessment is not a supplementary test but an integral part of the nature of osteoporosis. It is crucial for accurate diagnosis, independent prediction of fracture risk, monitoring treatment efficacy, understanding the mechanisms of bone strength, and guiding clinical decision-making. Measuring BMD alone is insufficient for a comprehensive assessment of osteoporosis; 3D bone microstructure must also be evaluated \cite{christiansen1991consensus}.

The assessment of three-dimensional bone microstructure is primarily based on morphological parameters.  Among these, trabecular bone morphology~\citep{parfitt1987bone}---quantified by trabecular thickness (Tb.Th), spacing (Tb.Sp), number (Tb.N), connectivity, bone volume fraction (BV/TV), and structure model index (SMI)---is central to osteoporosis diagnosis, fracture risk assessment, and bone microstructural analysis. However, the physiological scale of these parameters poses a fundamental challenge for imaging: the thickness of human bone trabeculae is only about 82–284 $\mu$m, while the spacing between trabeculae can range from 0.45 to 1.31 mm. To reliably segment and quantify structures at this scale in vivo, the spatial resolution of the imaging system must reach a range on the same order of magnitude as the dimensions of the bone trabeculae themselves—or even finer. To achieve this goal, numerous imaging systems and methods have been developed to resolve bone microstructure, such as high-resolution peripheral quantitative CT (HR-pQCT)~\cite{zhou2026time}, laboratory micro-CT~\cite{song2026review}, and two-dimensional histologic and histomorphometric studies \cite{DALLECARBONARE2005609,stout2019bone} have significantly improved the ability to assess three-dimensional bone microstructure morphology, it remains to be demonstrated whether more widely available routine clinical CT can extend imaging from the macroscopic to the mesoscopic scale and enable the quantitative assessment of bone microstructure parameters. 

The clinical CT and UHRCT (ultra-high-resolution computed tomography)~\citep{flohr2007novel} are well-suited for rapid imaging of macroscopic human anatomical structures with a wide field of view; however, they were not originally designed to clearly image small-scale microstructures. Due to their inherent characteristics, this can lead to biases in morphological quantitative analysis and precludes quantitative assessment.  SR$\mu$CT~\citep{grodzins1983optimum}, while capable of providing high-resolution references for microstructures, is not accessible to the general public. Can we use synchrotron micro-CT data—the “gold standard” for human tissue imaging—to enable the macroscopic-scale imaging system UHRCT to acquire mesoscale imaging capabilities without enhancing its physical imaging capabilities? How can we ensure the reliability of such results?

SR$\mu$CT and UHRCT are distinct physical imaging systems. These two imaging systems exhibit fundamental differences in terms of focal spot size, beam collimation, angular divergence, geometric magnification, detector element pitch, sampling density, detector PSF, quantum noise, motion artifacts and spatial coherence. Due to limitations in imaging capabilities, the reconstruction process of UHRCT relies on complex physical modeling and calibration. The most obvious differences between images of these two physical domains lie in spatial resolution, partial volume effects, noise, and artifacts. At $3.2~\mu$m, SR$\mu$CT resolves individual trabeculae with negligible partial volume effect and minimal noise, and is free of beam hardening and scatter. In contrast, clinical UHRCT at $100~\mu$m suffers from severe partial volume effect (bone/marrow mixing), strong quantum and electronic noise, detector blur, beam hardening, and Compton scatter. The resolution gap is $31.25\times$, far exceeding the high magnifications (such as 4x and 8x) used in the custom research datasets for super resolution method validation. The two domains also differ drastically in data volume and can only be coarsely paired---same anatomical site, different resolutions, no strict alignment.

\begin{figure}[t]
    \centering
    \includegraphics[width=0.9\linewidth]{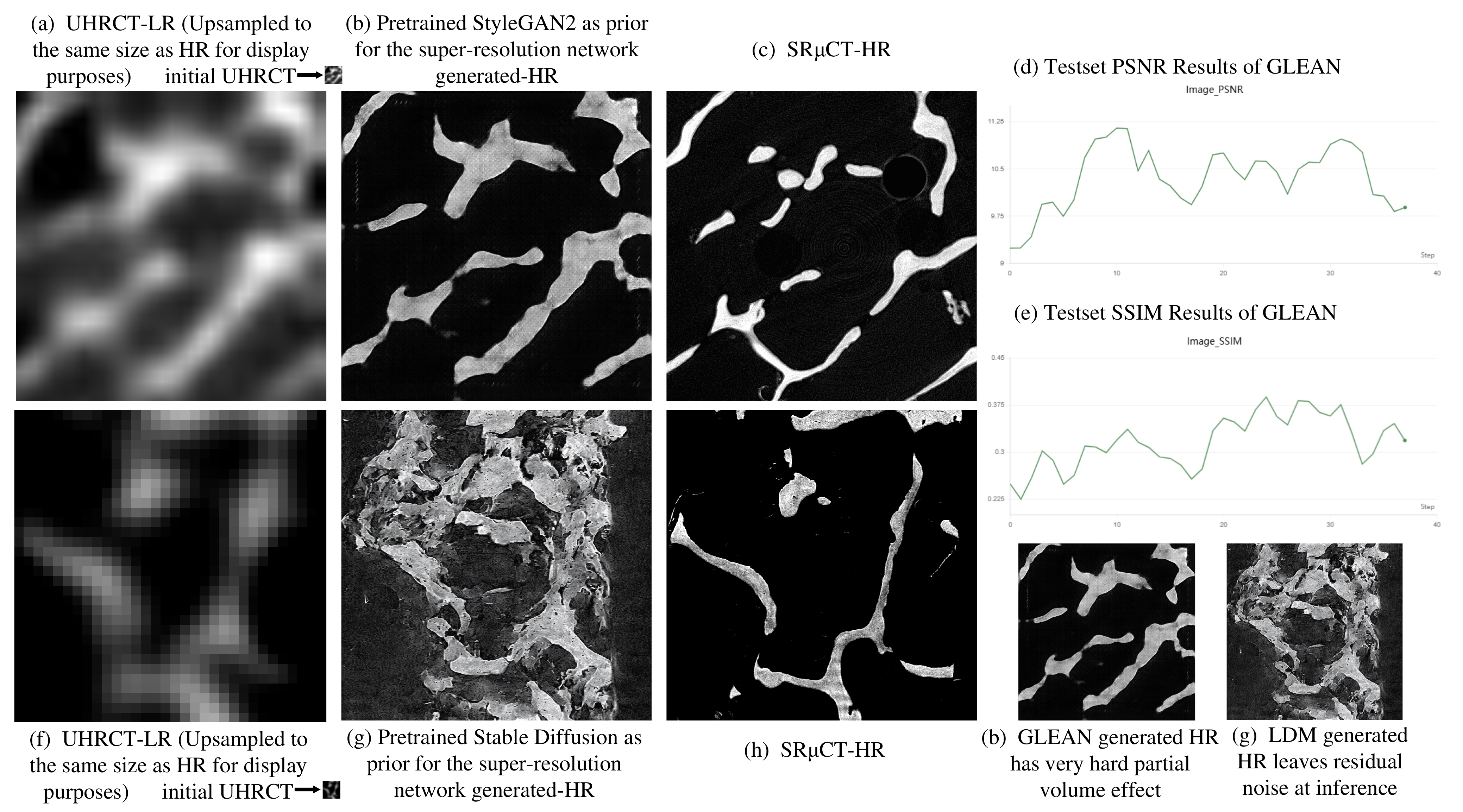}
    \caption{GLEAN fails to remove partial volume effects in clinical UHRCT as (a)-(e). Stable SR (using synthetic UHRCT data) leaves residual noise at inference as (f)-(h). For GLEAN, first, we performed large-scale pre-training of the generator using 309,482 high-resolution SR$\mu$CT images in StyleGAN2, and then used the generator as the prior for the super-resolution network.  (d) and (e) show the PSNR and SSIM results for the 37 CT images in the test set of GLEAN, respectively.}
    \label{fig:glean}
\end{figure}

Given the disparity in resolution, scale, as well as the scarcity of paired data, we consider pretraining on large-scale SR$\mu$CT data to learn prior knowledge about trabecular bone structure, and then using this knowledge to guide super-resolution processing. In recent years, GANs~\citep{goodfellow2014generative} and diffusion models~\citep{ho2020denoising,li2026back} achieve strong performance on natural image super-resolution. We adapt two representative large data pretrained methods to our task. For GAN-based high resolution (HR) priors, we adopt StyleGAN2~\citep{karras2020analyzing}. We then adapt GLEAN~\citep{9808408} to our task, which has been reported to support super-resolution of natural images at high magnifications. For diffusion-based priors, we adopt LDM~\citep{rombach2022high}. Then we adapt Stable SR~\citep{wang2024exploiting} to our task. However, they underperform: GLEAN fails to remove partial volume effects (Fig.~\ref{fig:glean} (b)), and Stable SR leaves residual noise at inference that contaminates reconstructed microstructures (Fig.~\ref{fig:glean} (g)). In a word, GLEAN (StyleGAN2) and Stable SR (LDM), based on pretrained priors, exploit large-scale high-resolution data via two-stage training. However, they capture source-specific texture statistics rather than invariant trabecular morphology, and hallucinate plausible but false trabeculae under coarse pairing. This is because our task is trustworthy microstructure inference, not diverse pixel generation.

The trustworthy macro-to-micro structural inference differs fundamentally from existing super-resolution. Existing networks are designed for a single objective---diverse detail generation for perceptual quality---and evaluated by pixel-level metrics (SSIM, PSNR, LPIPS). Our structural inference requires deterministic prediction of invariant microstructures from macro-scale low-resolution inputs, evaluated by morphological parameters for clinical trustworthiness, with minimal tolerance for hallucination. This demands a shift from pixel generation to topological inference.

How can we learn invariant trabecular morphology shared by both domains without large-scale pretraining and two-stage decoupling, reframing the objective as trustworthy deterministic structural inference? We propose structural dual network, a trustworthy, few-shot, end-to-end, lightweight approach. Its core idea is that forward physical degradation and inverse structural inference are not independent tasks, but a pair of mappings mutually constrained through shared morphological structures. The forward process $F$ models cross-modal physical degradation from microCT to UHRCT (micro-to-macro), including resolution loss, partial volume effects, noise, detector response differences, and artifacts. The inverse process $I$ performs corresponding inverse degradation operations and structural inference (macro-to-micro), recovering only the invariant trabecular morphology shared by both domains, without freely generating details. The two are coupled through structural duality constraints. We further propose a pyramid structural consistency discriminator operating on structural maps rather than raw images, extracting invariant structures, and four structural duality constraints to prevent hallucinated trabeculae. On the test set, SDN achieves morphological parameters largely consistent with SR$\mu$CT across 7 metrics (e.g., BV/TV 0.1163 vs.\ 0.1156, Tb.Th 10.3~$\mu$m vs.\ 7.3~$\mu$m), whereas clinical UHRCT alone yields large errors (BV/TV 0.3793, 228\% error; Tb.Th 357.2~$\mu$m, 4793\% error), with SSIM reaching 0.8. Cross-source generalization further proves that SDN can mine invariant structures and perform trustworthy inference. These validate the effectiveness and clinical trustworthiness of structural dual super-resolution for topological inference.

\textbf{Abbreviations}

BSRF = Beijing Synchrotron Radiation Facility, BV/TV = bone volume fraction, DA = degree of anisotropy, GAN = generative adversarial network, PSNR = peak signal-to-noise ratio, SDN = structural dual network, SMI = structure model index, SR$\mu$CT = synchrotron radiation micro-CT, SSIM = structural similarity index metric, SSRF = Shanghai Synchrotron Radiation Facility, Tb.Sp = trabecular separation, Tb.Th = trabecular thickness, UHRCT = ultra-high-resolution CT

\textbf{Summary}

A paradigm shift from pixel generation to topological inference is described for cross-physical-domain trabecular morphology learning from clinical UHRCT to SR$\mu$CT, realized by a few-shot structural dual network (SDN) that couples forward physical degradation with inverse structural inference through structural duality constraints; this shift is shown to be essential for efficiently solving the ill-posed super-resolution reconstruction problem and offers a new perspective for the research community.

\textbf{Key Points}
\begin{itemize}
 \item Topological inference super-resolution reframes 32× cross-domain super-resolution as deterministic structural inference rather than pixel generation.

 \item A few-shot SDN combining forward degradation, inverse inference, and structural duality constraints achieved morphological parameters largely consistent with SR$\mu$CT.

 \item Trained on 3.2 $\mu$m SSRF data, the model generalized to independent 3.25 $\mu$m BSRF data, supporting trustworthy cross-source structural inference.

\end{itemize}

\textbf{Keywords}

Trabecular microstructure, osteoporosis, ultra-high-resolution CT, synchrotron radiation micro-CT, super-resolution, topological inference, structural dual network, few-shot learning

\section{Materials and Methods}
\label{sec:method}

\subsection{Data}
\label{sec:data}

   \textbf{Synchrotron radiation microCT}: $3.2~\mu$m resolution, Shanghai Synchrotron Radiation Facility (SSRF), micro domain $X$.

   \textbf{Clinical UHRCT}: $100~\mu$m resolution, macro domain $Y$, with severe partial volume effects, noise, and artifacts.

 \textbf{Coarsely paired data}: from \textbf{6 patients}, totaling \textbf{1787 coarsely paired slices}. 5 patients for training, 1 patient for testing.

    \textbf{Generalization test set}: Beijing Synchrotron Radiation Facility (BSRF), same test patient, different resolution ($3.25~\mu$m isotropic), for cross-physical-domain generalization.

    \textbf{Ethics statement}: All data collection was approved by the Institutional Review Board and followed relevant guidelines for human sample research. Informed consent was obtained from all patients. The details of our dataset is in the appendix \ref{sec:appendix_dataset}.

\begin{figure}[t]
    \centering
    \includegraphics[width=0.9\linewidth]{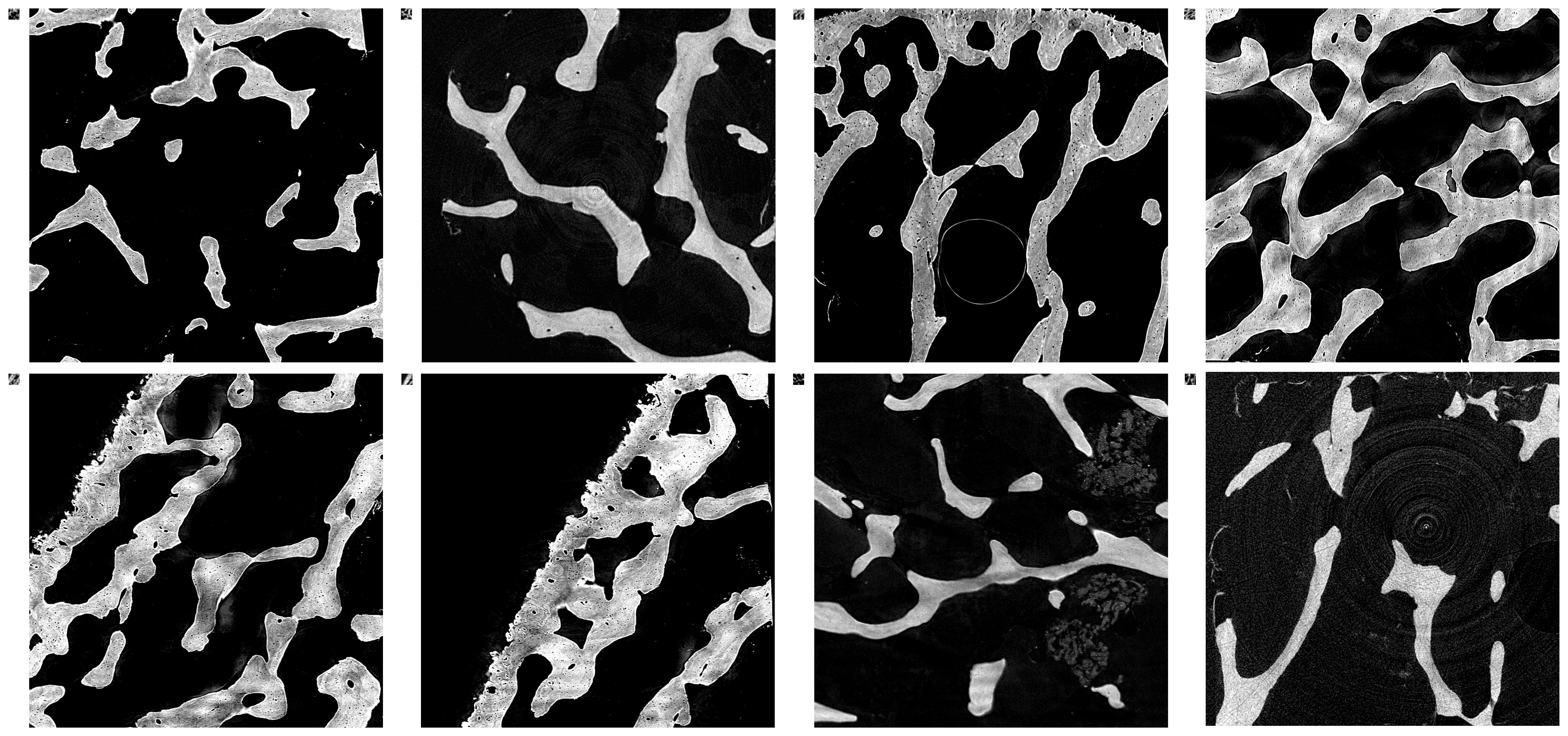}
    \caption{Examples of eight coarsely paired SR$\mu$CT--UHRCT images in the dataset. Each pair consists of a SR$\mu$CT slice (right large image) and the corresponding clinical UHRCT slice (left small image). The resolution gap is 31.25$\times$, and the two domains are not strictly aligned. The eighth pair (bottom right) is from BSRF, whereas the remaining seven pairs are from SSRF training data.}
    \label{fig:paired_examples_twosource}
\end{figure}

The dataset contains 1787 paired data from SSRF and 63 paired data from BSRF for generalization testing. Some paired data are shown at their original scales in Fig.~\ref{fig:paired_examples_twosource}; a version rescaled to a common size is provided in appendix \ref{sec:appendix_dataset} Fig.~\ref{fig:paired_examples}. A clearer illustration of the scale comparison is provided in appendix \ref{sec:appendix_dataset} Fig.~\ref{fig:size_comparison}.

\begin{table}[t]
\centering
\caption{Physical differences between synchrotron radiation microCT and clinical UHRCT.}
\label{tab:diff}

\begin{tabular}{lcc}
\toprule
Dimension & Synchrotron radiation microCT ($X$) & Clinical UHRCT ($Y$) \\
\midrule
Resolution & $3.2~\mu$m & $100~\mu$m \\
X-ray & Monochromatic & Polychromatic \\
Detector & Photon-counting & Energy-integrating \\
Partial volume effect & Negligible & Severe, bone/marrow mixing \\
Noise & Very low quantum noise & Quantum + electronic noise \\
Detector blur & Minimal & Significant \\
Beam hardening & None & Present \\
Scatter & None & Present (Compton) \\
Isotropy & Isotropic & Often anisotropic \\
\bottomrule
\end{tabular}
\end{table}

\subsection{Problem Setting}
\label{sec:problem}

Let two real physical domains be:
\begin{itemize}[label={}]
    \item $X$: synchrotron radiation microCT (SR$\mu$CT), $3.2~\mu$m resolution, monochromatic X-ray, photon-counting detector, negligible partial volume effect, minimal noise, no beam hardening or scatter artifacts.
    \item $Y$: clinical UHRCT, $100~\mu$m resolution, polychromatic X-ray, energy-integrating detector, severe partial volume effect, strong quantum and electronic noise, beam hardening, and scatter.
\end{itemize}
Data are \emph{coarsely paired}: same anatomical site, different resolutions, no strict alignment. The resolution gap is $31.25\times$, modeled as $32\times$ microstructure super-resolution. The physical differences between the two domains are summarized in Table~\ref{tab:diff}.

Goal: infer from $y \in Y$ the invariant trabecular morphology shared with $x \in X$.

We note that this goal fundamentally differs from conventional super-resolution. Conventional super-resolution aims to generate diverse, perceptually realistic details and is evaluated by pixel-level metrics (SSIM, PSNR, F1 Score). In contrast, our goal requires \emph{deterministic inference of invariant microstructures} from macro-scale low-resolution inputs, with minimal tolerance for hallucination. We therefore reframe the task as \emph{topological inference} rather than pixel generation.

\subsection{Paradigm: From Pixel Generation to Topological Inference}
\label{sec:paradigm}

We define \emph{topological inference super-resolution} as a new paradigm: given a macro-scale low-resolution input, deterministically infer the invariant topology and morphology of the underlying microstructure, without generating free-form texture details. Formally, the objective is:
\begin{equation}
\min \; d_{\text{struct}}\big(I(y), x\big), \quad \text{subject to} \quad I(y) \in \mathcal{M}(x),
\end{equation}
where $d_{\text{struct}}$ is a structural distance and $\mathcal{M}(x)$ denotes the set of microstructures sharing the same invariant topology as $x$. Unlike pixel generation, which minimizes $\|I(y) - x\|$ in image space, topological inference minimizes structural discrepancy, not pixel discrepancy.

This paradigm shift has three implications:
\begin{itemize}[label={}]
    \item \textbf{Objective}: from diverse detail generation to deterministic structural inference.
    \item \textbf{Evaluation}: from pixel-level metrics (SSIM/PSNR) to morphological parameters (Tb.Th, Tb.Sp, Tb.N, connectivity, BV/TV, SMI).
    \item \textbf{Constraint}: from single pixel-wise constraint to multi-scale structural duality, with minimal tolerance for hallucination.
\end{itemize}

\subsection{Realization: Structural Dual Network}
\label{sec:realization}

\begin{figure}[t]
    \centering
    \includegraphics[width=\linewidth]{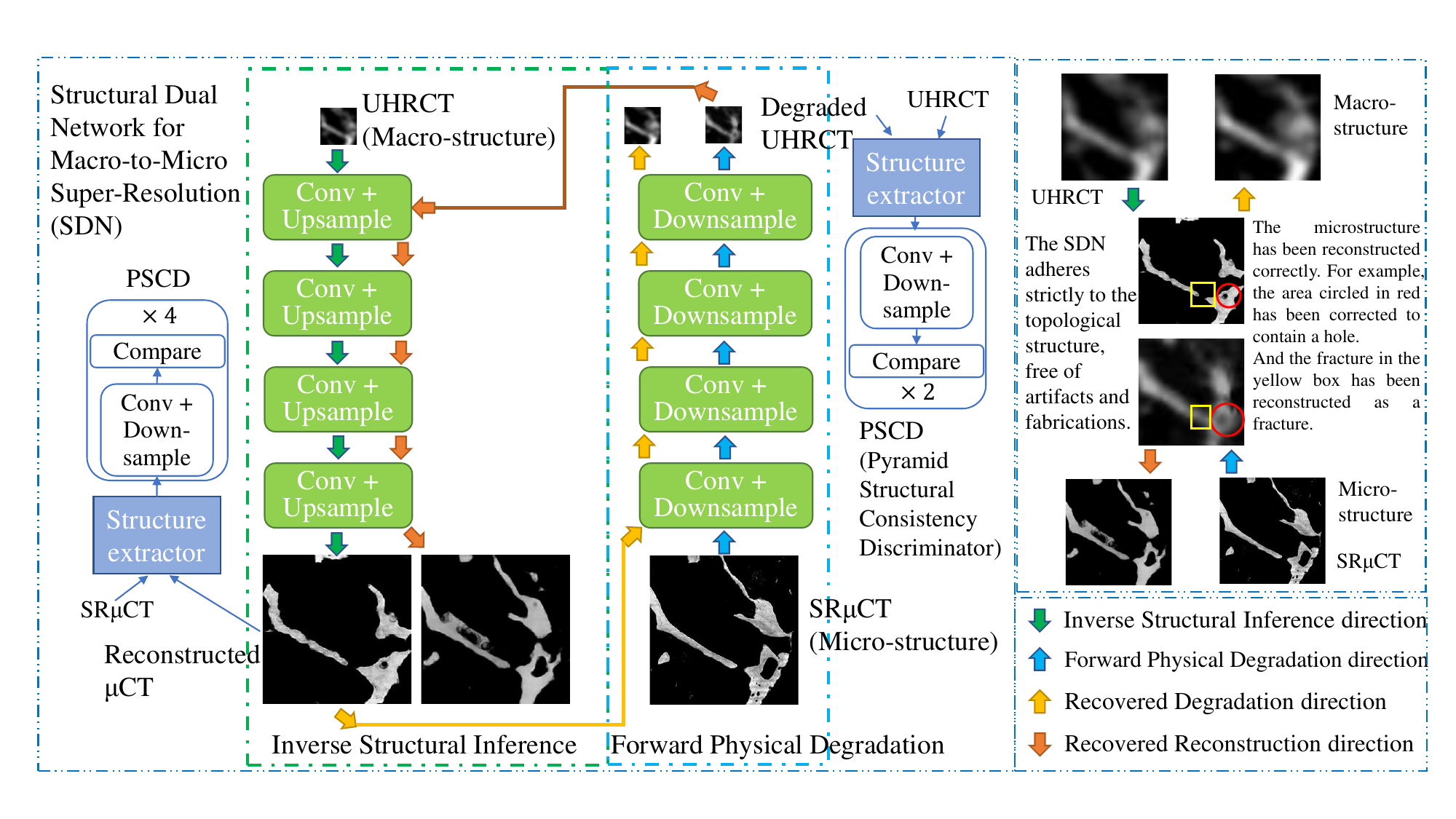}
    \caption{The framework of Structural Dual Network.}
    \label{fig:framework}
\end{figure}

We realize the topological inference paradigm via \emph{structural dual network}, illustrated in Fig.~\ref{fig:framework}, which couples two mappings:
\begin{equation}
F: X \rightarrow Y \quad \text{(forward physical degradation, micro-to-macro)},
\end{equation}
\begin{equation}
I: Y \rightarrow X \quad \text{(inverse structural inference, macro-to-micro)}.
\end{equation}
The two are coupled through \emph{structural duality constraints}:
\begin{equation}
d_{\text{struct}}\big(I(F(x)), x\big) \approx 0, \quad d_{\text{struct}}\big(F(I(y)), y\big) \approx 0,
\end{equation}
where $d_{\text{struct}}$ operates on low-frequency, gradient, SSIM, and frequency low-band representations, not pixel-level SSIM. This supplements pixel-wise cycle consistency with structural-level consistency, ensuring that the two mappings share the same invariant trabecular topology.

\noindent\textbf{Key insight}: topological inference does not require an explicit morphological parameter extractor during training. Instead, the structural duality constraints and the structural discriminator jointly enforce that the two mappings preserve invariant topology. Morphological parameters are used only for evaluation.

\subsection{Forward Physical Degradation: Micro-to-Macro}
\label{sec:forward}

The forward process $F$ models cross-modal physical degradation from microCT to UHRCT. We decompose $F$ into four physically interpretable components corresponding to the domain differences in Table~\ref{tab:diff}:
\begin{equation}
F = F_{\text{artifact}} \circ F_{\text{noise}} \circ F_{\text{PVE}} \circ F_{\text{res}}.
\end{equation}
\begin{enumerate}
    \item \textbf{Resolution loss $F_{\text{res}}$}: $3.2~\mu$m $\rightarrow$ $100~\mu$m, modeled as a learnable blur kernel followed by $32\times$ downsampling. The blur kernel approximates the detector point spread function of clinical UHRCT.
    \item \textbf{Partial volume effect $F_{\text{PVE}}$}: each voxel mixes trabeculae and marrow, biasing Tb.Th and Tb.Sp. We model this as structural unmixing and low-pass filtering, blurring bone/marrow boundaries.
    \item \textbf{Noise and detector response $F_{\text{noise}}$}: clinical UHRCT uses energy-integrating detectors with quantum noise, electronic noise, and detector blur; SR$\mu$CT uses photon-counting detectors with minimal noise. We model this as noise injection and detector response correction.
    \item \textbf{Artifacts $F_{\text{artifact}}$}: clinical UHRCT exhibits beam hardening and Compton scatter; SR$\mu$CT may have ring artifacts but far weaker. We model this as beam hardening residual, scatter superposition, and ring artifact suppression.
\end{enumerate}

\subsection{Inverse Structural Inference: Macro-to-Micro}
\label{sec:inverse}

The inverse process $I$ performs corresponding inverse degradation and structural inference. Each component corresponds to a physical degradation in $F$:
\begin{equation}
I = I_{\text{deartifact}} \circ I_{\text{denoise}} \circ I_{\text{dePVE}} \circ I_{\text{SR}}.
\end{equation}
\begin{enumerate}
    \item \textbf{Super-resolution/deconvolution $I_{\text{SR}}$}: inverse of $F_{\text{res}}$, deconvolving from $\approx 100~\mu$m to $3.2~\mu$m level, recovering trabecular edges blurred by the kernel.
    \item \textbf{Partial volume effect removal $I_{\text{dePVE}}$}: inverse of $F_{\text{PVE}}$, unmixing bone/marrow in each voxel to correct Tb.Th and Tb.Sp biases.
    \item \textbf{Denoising and detector response correction $I_{\text{denoise}}$}: inverse of $F_{\text{noise}}$, removing quantum and electronic noise, correcting detector blur, and restoring sharp edges.
    \item \textbf{Deartifact $I_{\text{deartifact}}$}: inverse of $F_{\text{artifact}}$, suppressing beam hardening, scatter, and ring artifacts, preventing them from being mistaken for trabeculae.
    \item \textbf{Invariant morphology inference}: recovering only shared trabecular morphology, without freely generating details.
\end{enumerate}
Since physical degradation is irreversible, $I$ is a \emph{structural-level inverse}, not a strict mathematical inverse of $F$. This is consistent with the topological inference paradigm: we do not aim to reconstruct the exact pixel-level microstructure, but to infer the invariant topology that both domains share.

\subsection{Pyramid Structural Consistency Discriminator}
\label{sec:discriminator}

The PSCD takes structural maps (edge, gradient, low-pass, low-frequency FFT) rather than raw images, avoiding modality style bias. This design is motivated by the domain differences in   Table~\ref{tab:diff}: raw images differ drastically in noise, contrast, and artifacts, but structural maps are largely invariant across domains. Multi-scale discrimination:
\begin{itemize}
    \item Low scale: macro trabecular layout.
    \item Mid scale: structural transitions.
    \item High scale: local morphology consistency.
\end{itemize}
Output is a patch-level score map, averaged over positions. By operating on structural maps, the discriminator enforces topological consistency rather than pixel-level realism, aligning with the topological inference paradigm.

\subsection{Loss Functions}
\label{sec:loss}

The F and I's loss:
\begin{equation}
\mathcal{L}_\text{F/I} = \mathcal{L}_{\text{adv}} + \lambda_1 \mathcal{L}_{\text{low}} + \lambda_2 \mathcal{L}_{\text{grad}} + \lambda_3 \mathcal{L}_{\text{ssim}} + \lambda_4 \mathcal{L}_{\text{freq}} + \lambda_5 \mathcal{L}_{\text{id}} + \lambda_6 \mathcal{L}_{\text{tv}}.
\end{equation}
The PSCD's loss:
\begin{equation}
\mathcal{L}_\text{PSCD} = \sum_{k=1}^K w_k \frac{1}{2} \Big[ \text{BCE}\big(D_k(\Phi(x)), 1\big) + \text{BCE}\big(D_k(\Phi(I(y))), 0\big) + \text{BCE}\big(D_k(\Phi(y)), 1\big) + \text{BCE}\big(D_k(\Phi(F(x))), 0\big) \Big],
\end{equation}
where $\Phi(\cdot)$ denotes structural map extraction, and $w_k$ is the scale weight.

\subsubsection{Pyramid Structural Consistency Adversarial Loss}
\begin{equation}
\mathcal{L}_{\text{adv}} = \sum_{k=1}^K w_k \Big[ \text{BCE}\big(D_k(\Phi(I(y))), 1\big) + \text{BCE}\big(D_k(\Phi(F(x))), 1\big) \Big].
\end{equation}

\subsubsection{Structural Duality Loss}
Let $\hat{x} = I(F(x))$, $\hat{y} = F(I(y))$.
\begin{align}
\mathcal{L}_{\text{low}} &= \| \text{LP}(\hat{x}) - \text{LP}(x) \|_1 + \| \text{LP}(\hat{y}) - \text{LP}(y) \|_1, \\
\mathcal{L}_{\text{grad}} &= \| \nabla \hat{x} - \nabla x \|_1 + \| \nabla \hat{y} - \nabla y \|_1, \\
\mathcal{L}_{\text{ssim}} &= \big(1 - \text{SSIM}(\hat{x}, x)\big) + \big(1 - \text{SSIM}(\hat{y}, y)\big), \\
\mathcal{L}_{\text{freq}} &= \| \mathcal{F}_{\text{low}}(\hat{x}) - \mathcal{F}_{\text{low}}(x) \|_1 + \| \mathcal{F}_{\text{low}}(\hat{y}) - \mathcal{F}_{\text{low}}(y) \|_1.
\end{align}
These four constraints jointly enforce topological consistency without requiring an explicit morphological parameter extractor. The structural distance $d_{\text{struct}}$ is implicitly defined by their combination.

\subsubsection{Regularization}
\begin{equation}
\mathcal{L}_{\text{id}} = \| I(x) - x \|_1 + \| F(y) - y \|_1,
\end{equation}
\begin{equation}
\mathcal{L}_{\text{tv}} = \text{TV}(I(y)) + \text{TV}(F(x)).
\end{equation}
$\mathcal{L}_{\text{id}}$ corresponds to $\lambda_5$, $\mathcal{L}_{\text{tv}}$ to $\lambda_6$. Recommended weights: $\lambda_{\text{adv}}=0.5$, $\lambda_1=1.0$, $\lambda_2=0.3$, $\lambda_3=0.5$, $\lambda_4=0.3$, $\lambda_5=0.5$, $\lambda_6=0.1$.

\subsection{Training Strategy}
\label{sec:training}

All components---the forward physical degradation network $F$, the inverse structural inference network $I$, and the pyramid structural consistency discriminator $\{D_k\}_{k=1}^K$---are trained jointly in an end-to-end manner, without pretraining or stage-wise decoupling. Training requires only \textbf{1787 coarsely paired samples} to converge. The training alternates between:
\begin{itemize}
    \item \textbf{PSCD update}: fix $I$ and $F$, update $\{D_k\}$ with $\mathcal{L}_\text{PSCD}$.
    \item \textbf{F and I update}: fix $\{D_k\}$, update $I$ and $F$ with $\mathcal{L}_\text{F/I}$.
\end{itemize}
The structural duality constraints act on both $I(F(x))$ and $F(I(y))$ paths simultaneously, coupling forward degradation and inverse inference. The model is simple, trained end-to-end, and converges with few samples.

\section{Results}
\label{sec:exp}

\subsection{Evaluation Metrics}
\label{sec:metrics}
In this work, evaluation is centered on \textbf{morphological parameters} rather than conventional pixel-level metrics. Specifically, we compute trabecular thickness (Tb.Th), trabecular separation (Tb.Sp), trabecular number (Tb.N), connectivity density (Conn.D), bone volume fraction (BV/TV), structure model index (SMI), and degree of anisotropy (DA). These parameters directly characterize trabecular bone morphology and are clinically meaningful for osteoporosis diagnosis and fracture risk assessment. These evaluations fulfill the \textbf{morphology learning} defined in this work: the model is required to infer invariant morphological structures shared across physical domains, not to generate perceptually plausible pixels. Therefore, morphological parameters serve as the primary evaluation criteria, while SSIM, PSNR, and Dice are reported only as auxiliary image-level references. By aligning evaluation with morphology, we ensure that the assessed quality reflects true structural fidelity rather than pixel-wise similarity. The detailed explanation and specific calculations of seven morphological parameters, please refer to the appendix \ref{app:morphology}.

\subsection{SSRF Results}
\label{sec:results}
On the SSRF test set, SDN achieves morphological parameters largely consistent with SR$\mu$CT across seven metrics, with SSIM reaching 0.8. The improvement is particularly pronounced for Tb.Th and Tb.Sp, which are most affected by partial volume effects in clinical UHRCT. 

\begin{figure}
    \centering
    \includegraphics[width=\linewidth]{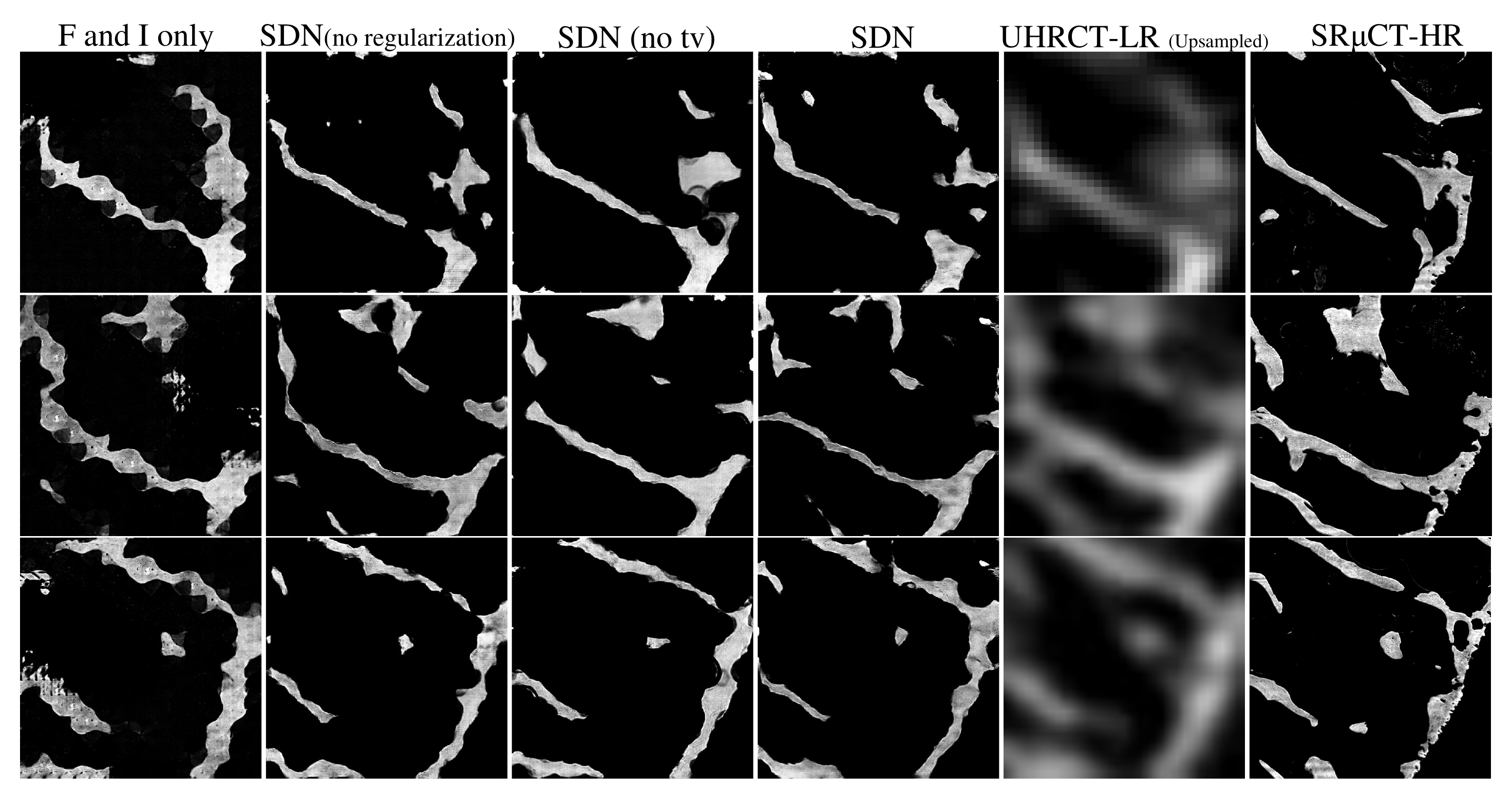}
    \caption{The results of ablation studies of Structural Dual Network.}
    \label{fig:ablation}
\end{figure}

\begin{table}[htbp]
\centering
\caption{Comparison of morphometric parameters on SSRF testset.}
\label{tab:six_groups}
\resizebox{\textwidth}{!}{
\begin{tabular}{lccccccc}
\toprule
%\rowcolor{gray!20}
Method & BV/TV & Tb.Th mean (mm) & Tb.Sp mean (mm) & Tb.N (1/mm) & SMI & DA & Conn.D (1/mm$^3$) \\
\midrule
UHRCT  & 0.3793 & 0.3572 & 0.3327 & 1.0619 & -0.7616 & 3.0564 & 1000.98 \\
SR$\mu$CT (GT)  & 0.1156 & 0.0073 & 0.0321 & 15.86 & -0.3444 & 334.49 & $3.05\times10^7$ \\
\rowcolor{green!10}
SDN & 0.1163 & 0.0103 & 0.0552 & 11.26 & -0.5053 & 446.54 & $3.05\times10^7$ \\
SDN (no tv)  & 0.1126 & 0.0100 & 0.0522 & 11.29 & -0.1594 & 382.83 & $3.05\times10^7$ \\
SDN (no reg) & 0.1472 & 0.0112 & 0.0473 & 13.16 & -0.4203 & 424.3 & $3.052\times10^7$ \\
F, I only & 0.1784 & 0.0123 & 0.0407 & 14.52 & -0.6024 & 377.3 & $3.052\times10^7$ \\

\bottomrule
\end{tabular}%
}
\end{table}

SDN achieves morphological parameters closely matching SR$\mu$CT. Given the 31.25$\times$ resolution gap, which we model as 32$\times$ super-resolution, discrepancies in Tb.Th (trabecular thickness) and Tb.Sp (trabecular separation) are unavoidable. Their slight deviations reflect the intrinsic information loss from 32$\times$ downsampling, which no method can fully recover.

We further compute PSNR, SSIM, and Dice (the F1 score in medical imaging), as shown in appendix~\ref{sec:appendix_results} Fig.~\ref{fig:metrics}. Compared with GLEAN on the test data, SDN achieves lower PSNR but significantly higher SSIM (0.8 vs.\ 0.3). This contrast reveals the fundamental difference between pixel-level recovery and structural inference. PSNR is sensitive to pixel-wise intensity differences; a lower PSNR indicates that SDN does not aim to reproduce exact pixel values, which are irreversibly lost in clinical UHRCT. In contrast, SSIM measures structural similarity, and the high SSIM demonstrates that SDN faithfully preserves trabecular topology. This confirms that SDN prioritizes structural inference and fidelity over pixel-level restoration, consistent with the topological inference paradigm.

Since GLEAN and Stable SR underperform when directly applied to our task, we further constructed large-scale synthetic LR-HR pairs by interpolating UHRCT data to train these two methods. The training set contains 52,874 LR-HR pairs, and the generative prior is pretrained on 309,482 high-resolution images. Test results are shown in Table~\ref{tab:morph_compare}. Compared with GLEAN and Stable SR trained on such large-scale synthetic interpolated data, SDN requires no synthetic data and achieves accurate structural inference using only 1787 coarsely paired real samples, demonstrating that it learns cross-domain invariant structures rather than synthetic degradation patterns. Despite being structurally simple, end-to-end, and trained on only 1787 coarsely paired samples, SDN outperforms these pretrained methods.

\begin{table}[htbp]
\centering
\caption{Comparison of morphometric parameters on interploated synthetic UHRCT testset: HR (SR$\mu$CT), LR (UHRCT), GLEAN and Stable SR. MAPE values are given in parentheses relative to HR. Dataset: 52874 paired UHRCT(contained interploated synthetic data)-SR$\mu$CT, pretrained on 309482 SR$\mu$CT.}
\label{tab:morph_compare}
\resizebox{\textwidth}{!}{%
\begin{tabular}{lcccc}
\toprule
Parameter & HR (SR$\mu$CT) & LR (UHRCT) & GLEAN~\citep{9808408} & Stable SR~\citep{wang2024exploiting} \\
\midrule
BV/TV            & 0.1284 & 0.37 & 0.1437 (MAPE 12.22\%) & 0.2406 (MAPE 86.72\%) \\
Tb.Th (mm)       & 0.0747 & 0.35 & 0.1152 (MAPE 57.82\%) & 0.0193 (MAPE 73.70\%) \\
Tb.Sp (mm)       & 0.5185 & 0.33 & 0.7128 (MAPE 40.77\%) & 0.0649 (MAPE 87.31\%) \\
Tb.N (mm$^{-1}$) & 1.7967 & 1.06 & 1.2647 (MAPE 27.76\%) & 13.2656 (MAPE 661.89\%) \\

\bottomrule
\end{tabular}%
}
\end{table}

\subsection{Ablation Studies}
\label{sec:ablation}

We conduct ablation studies on the SSRF testset (Fig.~\ref{fig:ablation}). Table~\ref{tab:six_groups} reports the morphological parameter values. Without structural duality constraints, the network (F and I) lacks explicit guidance to distinguish bone from marrow at the voxel level, and thus fails to correct the partial volume effect. Adding structural duality constraints significantly reduces these errors, as they enforce low-frequency and edge consistency across domains, effectively guiding the network to recover the underlying trabecular topology. Regularization further improves morphological accuracy by suppressing noise and preventing trivial mappings. The full SDN achieves the best performance across all metrics, confirming that each component contributes to the final morphological fidelity.

\begin{figure}
    \centering
    \includegraphics[width=\linewidth]{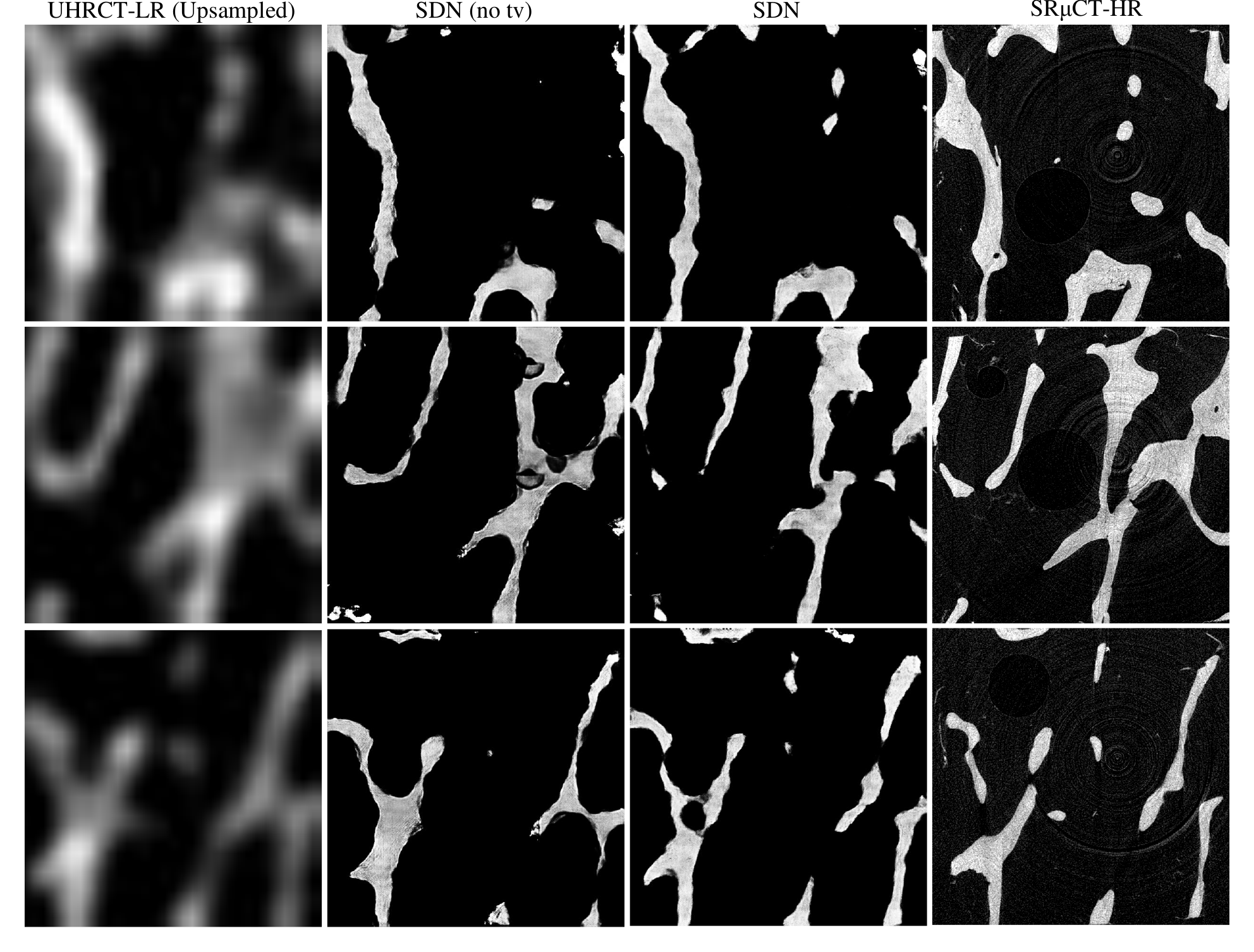}
    \caption{The results of cross-physical-domain generalization of Structural Dual Network.}
    \label{fig:bsrf}
\end{figure}

\subsection{Cross-Physical-Domain Generalization}
\label{sec:generalization}
The model is trained on $3.2~\mu$m SSRF data and generalizes well to $3.25~\mu$m isotropic BSRF data from an independent source. This validates that the model performs topological inference over pixel generation, learning invariant structures rather than source-specific styles. The BSRF SR$\mu$CT data contain prominent ring artifacts (last column, Fig. \ref{fig:bsrf}), and ring and motion artifacts were also present in the training data; nevertheless, SDN generated artifact-free bone microstructure (third column), demonstrating that SDN successfully extracted consistent structural information and removed X-ray micro-CT artifacts, thereby confirming that SDN achieved trustworthy cross-physical-domain trabecular morphology learning.

\begin{table}[htbp]
\centering
\caption{Comparison of morphometric parameters on BSRF testset for Cross-Physical-Domain Generalization.}
\label{tab:four_groups}
\resizebox{\textwidth}{!}{%
\begin{tabular}{lccccccccc}
\toprule
Method & BV/TV & Tb.Th mean  & Tb.Th std  & Tb.Sp mean & Tb.Sp std  & Tb.N (1/mm) & SMI & DA & Conn.D (1/mm$^3$) \\
\midrule
UHRCT& 0.2971 & 0.3235 & 0.0997 & 0.3832 & 0.2041 & 0.918 & 0.158 & 3.83 & 1000.9 \\
SR$\mu$CT  & 0.1425 & 0.0087 & 0.0052 & 0.0335 & 0.0255 & 16.41 & -0.270 & 285.4 & $2.91\times10^7$ \\
SDN & 0.1166 & 0.0105 & 0.0066 & 0.0535 & 0.0445 & 11.11 & -0.277 & 302.9 & $2.91\times10^7$ \\
SDN (no tv) & 0.0945 & 0.0091 & 0.0050 & 0.0539 & 0.0454 & 10.33 & -0.028 & 285.4 & $2.91\times10^7$ \\
\bottomrule
\end{tabular}%
}
\begin{flushleft}

\end{flushleft}
\end{table}

\subsection{Clinical Application: Whole Femoral Head Super-Resolution}
\label{sec:whole_femoral_head}
\begin{figure}
    \centering
    \includegraphics[width=\linewidth]{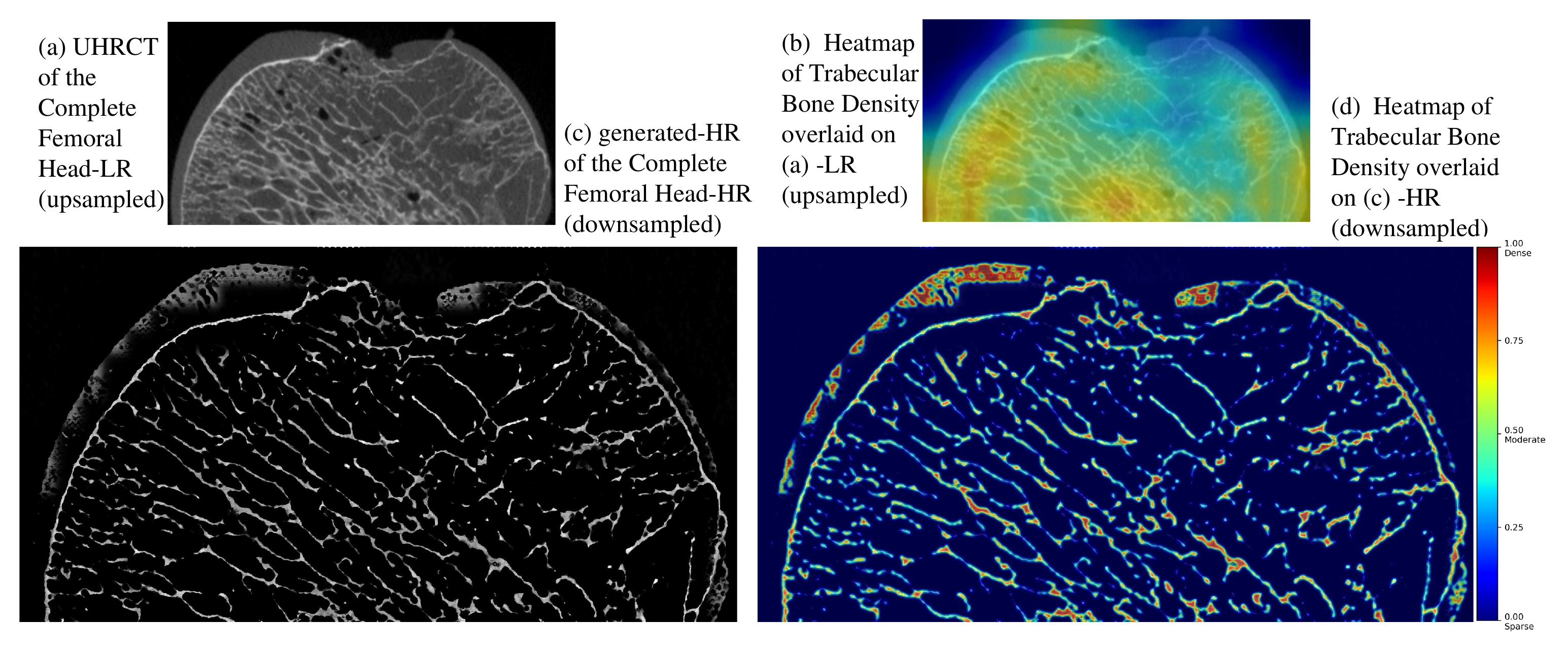}
    \caption{Clinical application: whole femoral head super-resolution. Partial images of the damaged bone tissue area have been excluded. (a) Input clinical UHRCT slice. (b) Heatmap on UHRCT, where most regions are incorrectly classified as high-density. (c) SDN SR reconstruction. (d) Heatmap on the reconstructed microCT, resolving individual trabeculae.}
    \label{fig:whole_femoral_head}
\end{figure}

Due to the limited field of view of synchrotron radiation sources, training data were acquired from \(4\times 4~\text{mm}^2\) trabecular bone specimens. The ultimate clinical goal, however, is to achieve super-resolution reconstruction of the complete femoral head. To demonstrate this clinical potential, we selected one slice from the complete femoral head UHRCT volume of the test set and applied SDN for super-resolution reconstruction, as shown in Fig.~\ref{fig:whole_femoral_head}.

Fig.~\ref{fig:whole_femoral_head}(b) and (d) show the local bone area fraction (B.Ar/T.Ar) heatmap overlaid on the input UHRCT and the reconstructed super-resolution microCT, respectively. The original UHRCT (Fig.~\ref{fig:whole_femoral_head}(b) classifies most regions as high-density areas, which is highly inaccurate due to severe partial volume effects. In contrast, the reconstructed heatmap (Fig.~\ref{fig:whole_femoral_head}(d) resolves individual trabeculae, where thicker trabeculae correspond to high-density regions, consistent with the microCT reference. This demonstrates that SDN enables accurate trabecular density quantification from clinical UHRCT, providing anatomical reference for precise screw placement in femoral head surgery.

\section{Discussion}
\label{sec:conclusion}
This study demonstrates that cross-physical-domain trabecular morphology learning from clinical UHRCT to SR$\mu$CT is feasible when super-resolution is reframed from pixel generation to topological inference. The central finding is that a 31.25× resolution gap—modeled as 32× microstructure super-resolution—can be addressed not by synthesizing plausible high-resolution pixels, but by deterministically inferring invariant trabecular structures that are evaluated by morphological parameters. This distinction is clinically consequential: in bone imaging, a visually plausible but false trabecula is not a benign artifact but a quantitative error that can propagate into fracture-risk assessment.

\textbf{Hallucination and trustworthiness.} Hallucination is a critical open problem in visual generation \citep{asadi2026mirage}. Generative models often fabricate plausible yet non-existent textures or structures \citep{chen2026survey}. Although such artifacts may be tolerable in natural images, in medical imaging any hallucinated structure can mislead diagnosis, quantification, and treatment planning. The medical domain therefore tolerates almost no hallucination and demands deterministic, structure-faithful reconstruction rather than free-form detail generation. This risk has direct economic and regulatory implications. A systematic review found that AI in radiology reduces costs only when accuracy matches human performance and may increase costs when specificity falls below human levels \citep{doi:10.1148/ryai.250090}; hallucinated structures would undermine precisely the diagnostic accuracy on which that value depends. The stakes are further clarified by the FDA authors’ report on image quality in the AI era, which warns that AI-based reconstruction and postprocessing can produce images that appear sharper, smoother, and more detailed while introducing new failure modes and exacerbating the disconnect between perceived image quality and diagnostic information content \citep{doi:10.1148/ryai.260100}. That report explicitly lists hallucinations and superresolution among the AI-specific failure modes requiring regulatory attention \citep{doi:10.1148/ryai.260100}. 

This concern is directly relevant to medical super-resolution, where pretrained-prior methods are particularly prone to such disconnect. Pretrained-prior methods such as GLEAN (StyleGAN2) and LDM (Stable Diffusion) exploit large-scale high-resolution data via two-stage training and achieve strong performance on natural image super-resolution. However, they capture source-specific texture statistics rather than invariant trabecular morphology, introduce textures irrelevant to the target domain under coarse pairing, and decouple the generative prior from the super-resolution objective, thereby hallucinating plausible but false trabeculae; they also fail to account for severe partial-volume effects and noise in clinical UHRCT. \citet{yazdan2026ghost} have begun to address hallucinations in multimodal LLMs, but hallucination in medical super-resolution remains largely unexplored. Our structural duality constraints—low-frequency, gradient, SSIM, and frequency low-band—were designed specifically to suppress this failure mode by enforcing consistency with forward physical degradation rather than encouraging free-form detail synthesis, so that the resulting morphological parameters reflect recoverable structure rather than generated appearance.

\textbf{Relation to existing medical image super-resolution.} Early medical image super-resolution methods use bicubic downsampling to synthesize low-resolution–high-resolution pairs, but real degradation is complex. Existing medical CT super-resolution methods likewise rely on synthetic degradation and pixel-wise losses \citep{peng2026super, hong2026enhancing, jhuboo2022state}, which in cross-modal, coarsely paired settings can hallucinate plausible but false trabeculae. Moreover, existing super-resolution typically targets the same imaging device with a resolution gap no more than 8× \citep{senck2025optimizing, frazer2024super, yu2022large}, far smaller than our 31.25× cross-device, cross-modal setting. Due to the limited number of human specimens and the scarcity of synchrotron facilities, large-scale imaging with the gold-standard SR$\mu$CT remains impractical, and research in this area is still in its early stages. Our results extend this landscape by showing that a few-shot structural dual network can operate under coarse pairing and severe domain discrepancy without large-scale pretraining.

\textbf{Trabecular morphology assessment and clinical relevance.} Following the WHO’s emphasis on the need to assess bone microstructural parameters, the field has increasingly adopted HR-pQCT \citep{boutroy2005vivo, manske2015human} for in vivo morphological assessment and microCT \citep{bouxsein2010guidelines} as an ex vivo reference for computing morphological parameters. However, HR-pQCT remains limited in anatomical coverage and availability, microCT is inherently ex vivo, and SR$\mu$CT—the gold standard for bone microarchitecture—cannot be applied in vivo, whereas clinical UHRCT is accessible but lacks the precision required for quantitative assessment. This gap motivates the problem we address: we are the first to formulate cross-physical-domain trabecular morphology learning from clinical UHRCT to SR$\mu$CT, with morphological parameters as the primary evaluation. The central difficulty is that these two domains differ fundamentally in resolution, partial-volume effect, noise, and artifacts. 

To address this challenge, we propose topological inference super-resolution, a paradigm shifting from pixel generation to deterministic structural inference, and realize it via structural dual super-resolution—coupling forward physical degradation (micro-to-macro) with inverse structural inference (macro-to-micro) through structural duality constraints. Specifically, we introduce a trustworthy, simple, end-to-end, lightweight, few-shot structural dual network with a pyramid structural consistency discriminator and four structural duality constraints. For evaluation, we establish a trustworthy evaluation protocol using morphological parameters rather than SSIM/PSNR, and demonstrate cross-physical-domain generalization from 3.2 $\mu$m SSRF to 3.25 $\mu$m isotropic BSRF data. Under this evaluation protocol, the model outperforms GAN and diffusion methods that rely on large-scale pretrained models, validating the effectiveness and clinical trustworthiness of structural dual super-resolution for topological inference. Cross-source generalization further indicates that SDN can mine invariant structures and perform trustworthy inference rather than overfitting to limited single-source data. Collectively, this study represents a step toward accurate in vivo quantification of bone microstructural parameters.

\textbf{Limitations and future directions.} Several limitations remain. First, morphology learning was trained on 2D slices yet evaluated on 3D morphological parameters; although this 2D-to-3D structural inference already yields metrics closely matching SR$\mu$CT, future work may explore 3D training to further improve morphological accuracy. Second, the available paired data remain scarce and coarsely aligned, reflecting the practical constraints of synchrotron facilities and human specimens; larger and more diverse paired datasets would strengthen generalizability. Third, the present evaluation covers two synchrotron sources; broader validation across scanners, acquisition protocols, and anatomical sites is needed before clinical deployment. Finally, with the completion and operation of HEPS (High Energy Photon Source), a more powerful fourth-generation synchrotron light source, larger-field-of-view and higher-resolution imaging will become possible, which should further improve the accuracy of morphological quantification in bone specimens and, ultimately, help bring bone microstructural assessment into routine clinical practice for fracture risk assessment.

\subsection*{AI use statement}

%(This section is \textbf{required} and does not count toward the page limit.)

In this work, we used generative AI tools for polish writing.
We have not used generative AI tools for other tasks with required disclosure,
and the rest of the required disclosure tasks are not applicable to this work.
We have reviewed all AI-assisted work. We take responsibility for the final content of this work,
including text, claims produced with the aid of generative AI.

\subsection*{Ethics statement}

%(This section is \textbf{recommended} and does not count toward the page limit.)

Compliance with Ethical Standards. This research was conducted in full accordance with the principles of the Declaration of Helsinki and all applicable national laws, regulations, and institutional policies governing research involving human participants. The study protocol was reviewed and approved by the Biomedical Ethics Committee of the authors' Institution. All human participants provided informed consent prior to their involvement in the study, and their data were anonymized to protect privacy. The authors have designed this study to avoid any foreseeable risks of harm, discrimination, or privacy violation.

\subsection*{Reproducibility statement}

%(This section is \textbf{recommended} and does not count toward the page limit.)
We are committed to ensuring the reproducibility of our work. To this end, we provide the following resources and details:

The details of data processing steps are specified in Appendix \ref{sec:appendix_dataset} 

The evaluation metrics are detailed definitions and calculation procedures for the proposed evaluation metrics are specified in Appendix \ref{app:morphology}.

Results from large-scale pre-training experiments using datasets processed with interpolation synthesis are provided in the Appendix \ref{sec:appendix_results}.

%\subsubsection*{Author Contributions}
%If you'd like to, you may include  a section for author contributions as is done
%in many journals. This is optional and at the discretion of the authors.

\subsubsection*{Acknowledgments}
%Use unnumbered third level headings for the acknowledgments. All
%acknowledgments, including those to funding agencies, go at the end of the paper.
We thank the 4W1A X-ray Imaging Beamline at the Beijing Synchrotron Radiation Facility (BSRF) and the BL13HB Beamline at the Shanghai Synchrotron Radiation Facility (SSRF) for their support in data acquisition. %We also thank Yuxuan Cao for providing the experimental results of Stable SR on synthetic data.

\bibliography{bibtex}
\bibliographystyle{iclr2027_conference}

\appendix

\let\origsubsection\subsection
\renewcommand{\subsection}[1]{%
  \origsubsection{#1}%
  \addcontentsline{app}{subsection}{\protect\numberline{\thesubsection}#1}%
}

\listofappendix
%\appsection{Appendix}
\label{sec:appendix}
%\subsection{Dataset}

\appsection{Data Description}
\label{sec:appendix_dataset}
This study constructs a paired imaging dataset of synchrotron radiation microCT (SR$\mu$CT) and clinical CT for human osteoporotic bone samples. The dataset includes discarded bone samples, high-resolution SR$\mu$CT images, clinical CT images, and registered cross-modal paired data.

\subsection{Sample Collection and Ethical Approval}

Discarded human femoral head samples after femoral head replacement were collected with approval from the hospital ethics committee. Patient informed consent was waived under the clause of ``using discarded medical samples without identifiable personal information.''

Inclusion criteria: (1) elderly patients aged over 65 years; (2) clinically and radiologically diagnosed with primary osteoporosis; (3) fragility fracture of the femoral neck caused by osteoporosis requiring surgical resection of the hip joint. Exclusion criteria: (1) severe degenerative changes causing bone structural destruction; (2) improper sample preservation resulting in putrefaction; (3) bone tumors, infection, secondary osteoporosis, or other metabolic bone diseases that significantly affect bone microstructure.

After collection, bone samples were immediately fixed in 4\% paraformaldehyde for 24--48 h and then stored in 70\% ethanol at 4\,$^\circ$C. During scanning, samples were kept moist to avoid trabecular microcracks and dimensional shrinkage caused by drying. For femoral head samples, after embedding in low-viscosity epoxy resin or PMMA, a low-speed precision cutter (e.g., a diamond wire saw) was used to cut the samples into bone strips with a cross-section of approximately 4~mm $\times$ 4~mm. The length was determined according to the SR$\mu$CT beamline field of view and sample holder size, typically 10--20~mm, ensuring that the region of interest was within the beamline field of view. During cutting, saline or ethanol was continuously applied for cooling and lubrication to avoid thermal damage and mechanical destruction. After cutting, the integrity of the bone strips was examined, and regions with intact trabecular structure were retained for imaging. If necessary, micro-markers were implanted into the bone strips to provide spatial corresponding points for subsequent cross-modal image registration.

\subsection{Synchrotron Radiation CT Imaging}

\textbf{Shanghai Synchrotron Radiation Facility BL13HB Beamline}

High-resolution micro-CT imaging of bone samples was performed at the Shanghai Synchrotron Radiation Facility (SSRF) BL13HB beamline. This beamline is dedicated to X-ray imaging and biomedical applications, with a maximum spatial resolution of 0.8~$\mu$m and an adjustable energy range of 8--40~keV, meeting the requirements for micron-scale three-dimensional imaging of trabecular bone. The cut bone strips with a cross-section of approximately 4~mm $\times$ 4~mm can be fully placed within the field of view of this beamline.

Imaging parameters were set as follows:

\begin{itemize}
    \item Imaging mode: absorption contrast imaging + phase contrast imaging;
    \item X-ray energy: 25--35~keV, optimized according to sample thickness;
    \item Reconstructed voxel size: 3.2~$\mu$m;
    \item Number of projections: 1200--1800 over 360$^\circ$ rotation;
    \item Exposure time: 0.5--2~s/projection, adjusted according to photon flux;
    \item Scan time: approximately 1--2~h/sample.
\end{itemize}

\textbf{Beijing Synchrotron Radiation Facility 4W1A X-ray Imaging Beamline}

In addition, for cross-source generalization the supplementary SR$\mu$CT imaging was performed at the Beijing Synchrotron Radiation Facility (BSRF) 4W1A X-ray Imaging Beamline. This beamline adopts an absorption/phase contrast imaging mode similar to that of the SSRF BL13HB beamline, with an energy range of approximately 6--26~keV and a beam spot size of approximately 10~$\times$~5~mm$^2$ at spatial resolution of ~4~$\sim4\mu$m. Bone strips with a cross-section of 4~mm $\times$ 4~mm can be fully covered. Limited by the light source and detector of the station, the effective spatial resolution is approximately 10~$\mu$m, and the reconstructed voxel size can be set to 3.25~$\mu$m. The number of projections, exposure time, and scan time were kept consistent with or similar to the above SSRF BL13HB protocol.

Spot parameters for micrometer-scale imaging:
\begin{itemize}
    \item Energy range: 8–26 keV;
    \item Photon flux at the sample (phs/s): $\sim10^{12}$ @ 8 keV;
    \item At spatial resolutions of 10~$\mu$m, 4~$\mu$m, 2~$\mu$m, and 1~$\mu$m, the spot sizes (H~$\times$V) are 13 mm~$\times$~13 mm, 10 mm~$\times$~5 mm, 5 mm~$\times$~2.5 mm, and 2 mm~$\times$~1 mm, respectively.
\end{itemize}

\subsection{Clinical CT Imaging}

The same batch of samples was scanned by clinical CT using an Ultra 3D UHRCT (Beijing Langshi Instrument Co., Ltd., Beijing, China). The scanning parameters were set as tube voltage 100--120~kV and tube current-time product 8--10~mAs. Two parameter settings were used:

\begin{itemize}
    \item Protocol A: field of view 8~cm $\times$ 8~cm, slice thickness and spacing both 0.15~mm;
    \item Protocol B: field of view 7~cm $\times$ 4~cm, slice thickness and spacing both 0.1~mm.
\end{itemize}

Based on the two protocols above, clinical CT images with two resolutions were reconstructed.

\subsection{Image Registration and Dataset Construction}

Due to significant differences between SR$\mu$CT and clinical CT in resolution, imaging coordinate system, and physical contrast mechanism, image registration is required. The registration workflow includes: coarse registration based on micro-markers implanted during sample preparation or anatomical landmarks; cross-modal rigid registration using mutual information or normalized mutual information as the similarity measure; and cropping to the common field of view of the two imaging modalities after registration. The recommended scanning order is to complete clinical CT scanning first, followed by SR$\mu$CT scanning, and to use the same fixation device as much as possible to maintain consistent spatial positions of the samples. During registration, the transformation matrix, marker coordinates, and registered image data were saved.

\subsection{Dataset Composition}

A paired SR$\mu$CT and clinical UHRCT imaging dataset of bone samples was established. Each sample includes:

\begin{itemize}
    \item SR$\mu$CT raw data (SSRF reconstructed voxel size 3.2~$\mu$m; BSRF reconstructed voxel size 3.25~$\mu$m);
    \item Clinical CT raw data (slice thickness/spacing 0.15~mm or 0.1~mm; reconstructed voxel size approximately 0.1~mm);
    \item Registered and cropped paired SR$\mu$CT and clinical CT data;
    \item Image registration transformation matrix and marker coordinates;
    \item Sample metadata, including age, sex, bone mineral density, clinical diagnosis, scanning parameters, and ethical approval number.
\end{itemize}

\begin{figure}[t]
    \centering
    \includegraphics[width=0.9\linewidth]{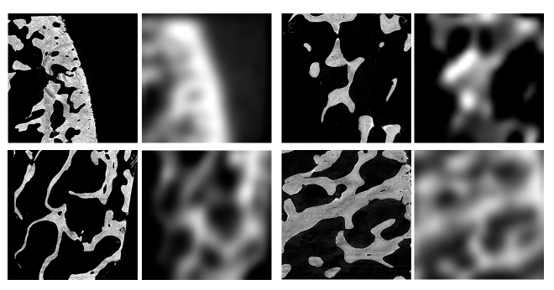}
    \caption{Examples of four coarsely paired SR$\mu$CT--UHRCT images in our dataset. Each pair consists of a SR$\mu$CT slice (left) and the corresponding clinical UHRCT slice (right) at the same anatomical site. The resolution gap is 31.25$\times$, and the two domains are not strictly aligned. Both images are displayed at the same physical size to highlight the 31.25$\times$ resolution gap.}
    \label{fig:paired_examples}
\end{figure}

\begin{figure}
    \centering
    \includegraphics[width=0.9\linewidth]{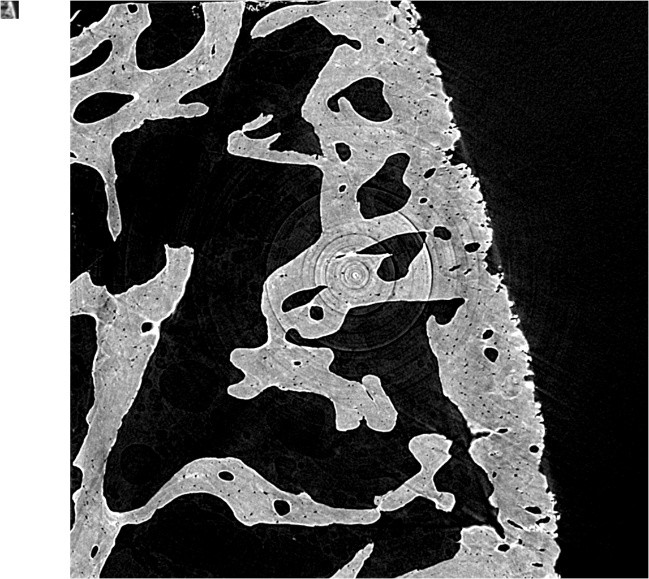}
    \caption{Real-size comparison between clinical UHRCT and SR$\mu$CT. The left panel shows UHRCT at 100~$\mu$m resolution, where individual trabeculae are indistinguishable due to severe partial volume effects. The right panel shows SR$\mu$CT at 3.2~$\mu$m resolution, resolving individual trabecular structures.}
    \label{fig:size_comparison}
\end{figure}

The dataset contains 1787 paired data from SSRF and 63 paired data from BSRF for generalization testing. Some paired data are illustrated in Fig.~\ref{fig:paired_examples}. The initial size of UHRCT and SR$\mu$CT is shown in Fig.~\ref{fig:size_comparison}.

\appsection{Morphological Parameters}
\label{app:morphology}

This appendix provides detailed definitions, calculation methods, and clinical significance of the morphological parameters used for evaluation. All parameters are computed on binarized 3D trabecular bone images.

\subsection{Degree of Anisotropy (DA)}
\label{app:da}

\textbf{Definition.} Degree of anisotropy quantifies the directional dependence of trabecular bone architecture---the extent to which trabeculae are preferentially oriented along specific directions rather than randomly distributed.

\textbf{Calculation.} DA is computed using the mean intercept length (MIL) method. Parallel test lines are traced in multiple directions across the binarized image, and the mean intercept length $\mathrm{MIL}(\omega)$ for each direction $\omega$ is obtained as
\begin{equation}
\mathrm{MIL}(\omega) = \frac{L}{I(\omega)},
\end{equation}
where $L$ is the total line length and $I(\omega)$ is the number of intersections with the bone--marrow interface. The MIL values define an ellipsoid, and DA is expressed as the ratio of the maximum to minimum eigenvalues of the fabric tensor:
\begin{equation}
\mathrm{DA} = 1 - \frac{\lambda_1}{\lambda_3},
\end{equation}
where $\lambda_1$ and $\lambda_3$ are the smallest and largest eigenvalues, respectively. DA ranges from 0 (fully isotropic) to 1 (fully anisotropic).

\textbf{Clinical significance.} Trabecular bone adapts its orientation to principal loading directions. In osteoporosis, the preferential orientation is lost and DA decreases, reducing load-bearing efficiency. DA is an independent predictor of bone stiffness and fracture risk.

\subsection{Bone Volume Fraction (BV/TV)}
\label{app:bvtv}

\textbf{Definition.} BV/TV is the ratio of mineralized bone volume (BV) to the total volume of interest (TV), expressed as a percentage.

\textbf{Calculation.}
\begin{equation}
\mathrm{BV/TV} = \frac{\mathrm{BV}}{\mathrm{TV}} \times 100\%.
\end{equation}
In practice, BV is the number of bone voxels after binarization, and TV is the total number of voxels within the region of interest.

\textbf{Clinical significance.} BV/TV directly reflects bone mass per unit volume and is the primary predictor of trabecular mechanical properties, accounting for approximately 87\% of trabecular stiffness. When BV/TV falls below $\sim$15\%, structural integrity is critically compromised and fracture risk rises sharply.

\subsection{Trabecular Thickness (Tb.Th)}
\label{app:tbth}

\textbf{Definition.} Tb.Th is the average thickness of individual trabecular struts.

\textbf{Calculation.} Tb.Th is computed by the maximal sphere fitting method: for each bone voxel, the diameter of the largest sphere that can be inscribed within the bone structure is determined, and these diameters are averaged over all bone voxels:
\begin{equation}
\mathrm{Tb.Th} = \frac{1}{\mathrm{Vol}(\Omega)} \int_{\Omega} \tau(\mathbf{x}) \, d^3\mathbf{x},
\end{equation}
where $\tau(\mathbf{x})$ is the local thickness at voxel $\mathbf{x}$ and $\Omega$ is the bone phase. Alternatively, under a plate-model assumption,
\begin{equation}
\mathrm{Tb.Th} = \frac{2}{\mathrm{BS/BV}},
\end{equation}
where $\mathrm{BS/BV}$ is the bone surface-to-volume ratio.

\textbf{Clinical significance.} Tb.Th reflects the thickness of individual trabecular elements. In osteoporosis, trabecular thinning occurs as a result of bone resorption, reducing load-bearing capacity.

\subsection{Trabecular Separation (Tb.Sp)}
\label{app:tbsp}

\textbf{Definition.} Tb.Sp is the average width of the marrow cavities between trabeculae, representing the porosity of the trabecular network.

\textbf{Calculation.} Tb.Sp is computed using the same sphere-fitting method as Tb.Th but applied to the marrow (background) phase:
\begin{equation}
\mathrm{Tb.Sp} = \frac{1}{\mathrm{Vol}(\Omega_{\text{marrow}})} \int_{\Omega_{\text{marrow}}} \tau(\mathbf{x}) \, d^3\mathbf{x}.
\end{equation}
Under the plate model, it can also be derived as
\begin{equation}
\mathrm{Tb.Sp} = \frac{2(1 - \mathrm{BV/TV})}{\mathrm{BS/TV}},
\end{equation}
or equivalently,
\begin{equation}
\mathrm{Tb.Sp} = \frac{1}{\mathrm{Tb.N}} - \mathrm{Tb.Th}.
\end{equation}

\textbf{Clinical significance.} Tb.Sp measures the widening of marrow spaces due to trabecular resorption. Increased Tb.Sp indicates a sparser, more porous trabecular network, which is characteristic of osteoporotic bone.

\subsection{Trabecular Number (Tb.N)}
\label{app:tbn}

\textbf{Definition.} Tb.N is the average number of trabeculae encountered per unit length along a linear path through the network.

\textbf{Calculation.}
\begin{equation}
\mathrm{Tb.N} = \frac{\mathrm{BV/TV}}{\mathrm{Tb.Th}},
\end{equation}
with units of $\mathrm{mm}^{-1}$.

\textbf{Clinical significance.} Tb.N reflects the density of the trabecular network. Reduced Tb.N indicates trabecular loss and network disruption. Notably, Tb.N reduction often occurs earlier than Tb.Th reduction in osteoporosis, making it a sensitive early indicator of microstructural deterioration.

\subsection{Connectivity Density (Conn.D)}
\label{app:connd}

\textbf{Definition.} Conn.D quantifies the number of redundant connections per unit volume in the trabecular network.

\textbf{Calculation.} Conn.D is derived from the Euler characteristic $\chi$ of the binarized bone structure. The connectivity $\beta_1$ is
\begin{equation}
\beta_1 = 1 - (\chi + \Delta\chi),
\end{equation}
where $\Delta\chi$ is a boundary correction term. Connectivity density is then
\begin{equation}
\mathrm{Conn.D} = \frac{\beta_1}{\mathrm{TV}},
\end{equation}
with units of $\mathrm{mm}^{-3}$.

\textbf{Clinical significance.} Connectivity describes the maximum number of trabecular connections that can be removed without disintegrating the structure into separate parts. Loss of connectivity indicates that the trabecular network is fragmenting, which severely compromises mechanical integrity. Conn.D is an independent predictor of bone strength.

\subsection{Structure Model Index (SMI)}
\label{app:smi}

\textbf{Definition.} SMI quantifies the characteristic geometry of trabeculae---whether they are predominantly plate-like (SMI $=0$) or rod-like (SMI $=3$).

\textbf{Calculation.} SMI is computed by a differential analysis of the triangulated bone surface. The bone surface is dilated infinitesimally in the normal direction, and the change in surface area $\mathrm{dBS}/\mathrm{dr}$ is determined:
\begin{equation}
\mathrm{SMI} = 6 \cdot \frac{\mathrm{BV} \cdot \frac{\mathrm{dBS}}{\mathrm{dr}}}{\mathrm{BS}^2}.
\end{equation}
Ideal flat plates yield SMI $=0$ (no surface change with dilation); ideal cylindrical rods yield SMI $=3$ (linear surface increase). Concave surfaces produce negative SMI values.

\textbf{Clinical significance.} Normal trabecular bone is predominantly plate-like (SMI close to 0). In osteoporosis, plates perforate and convert to rods, increasing SMI. This architectural conversion reduces compressive strength and is a hallmark of osteoporotic microstructural degradation.

\subsection{Summary}
\label{app:summary}

Table~\ref{tab:appendix_morph} summarizes the definitions, units, and osteoporotic changes of all morphological parameters.

\begin{table}[htbp]
\centering
\caption{Summary of morphological parameters.}
\label{tab:appendix_morph}
\begin{tabular}{llll}
\toprule
Parameter & Unit & Definition & Osteoporotic Change \\
\midrule
DA & --- & Degree of anisotropy (0 = isotropic, 1 = anisotropic) & Decreases \\
BV/TV & \% & Bone volume fraction & Decreases \\
Tb.Th & $\mu$m & Trabecular thickness & Decreases \\
Tb.Sp & $\mu$m & Trabecular separation & Increases \\
Tb.N & mm$^{-1}$ & Trabecular number & Decreases \\
Conn.D & mm$^{-3}$ & Connectivity density & Decreases \\
SMI & --- & Structure model index (0 = plate, 3 = rod) & Increases \\
\bottomrule
\end{tabular}
\end{table}

\appsection{Supplementary Results}
\label{sec:appendix_results}

\begin{figure}[t]
    \centering
    \includegraphics[width=\linewidth]{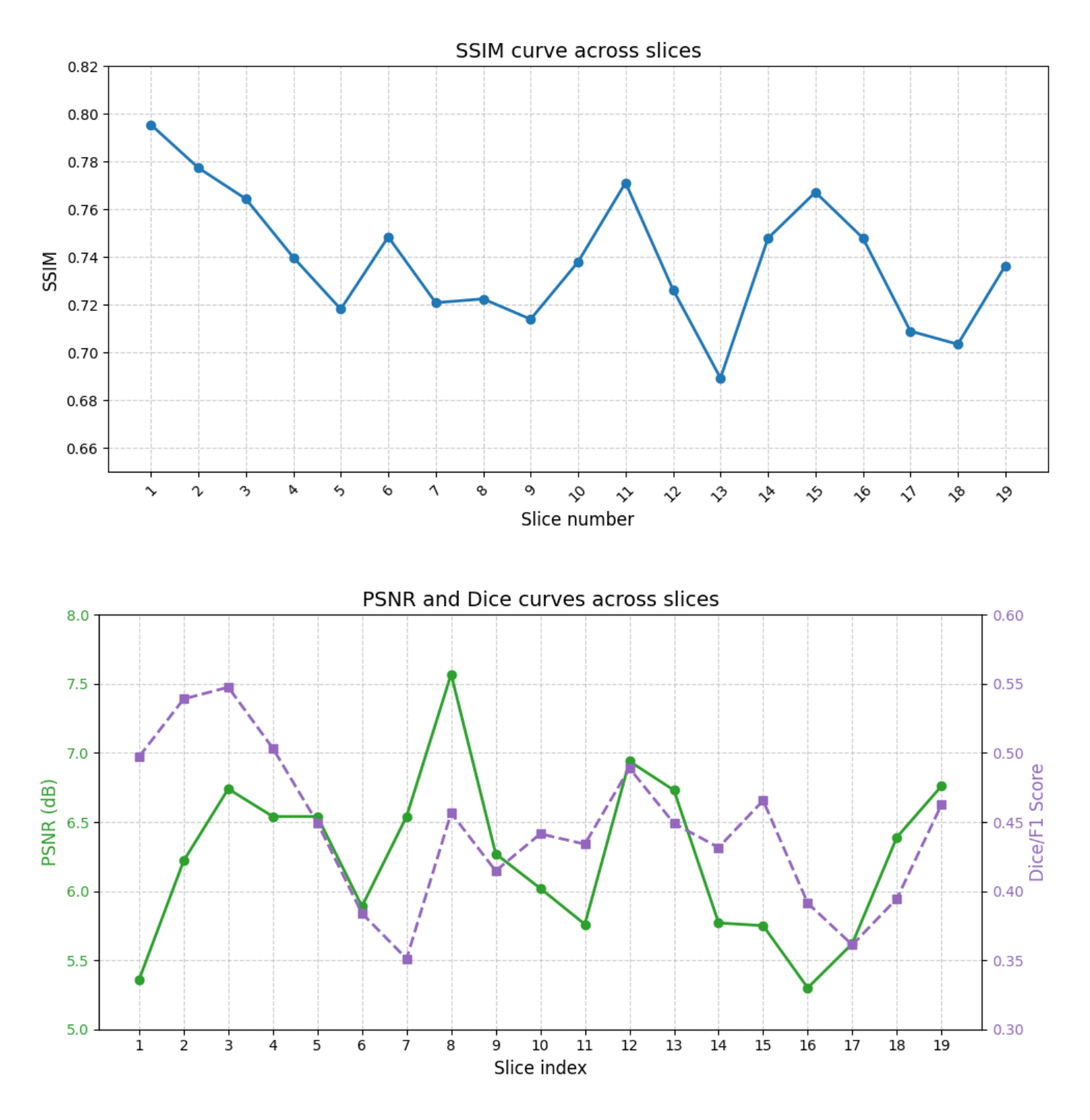}
    \caption{Image-level evaluation metrics. SDN achieves lower PSNR but significantly higher SSIM than GLEAN, indicating that SDN focuses on structural inference rather than pixel-level restoration.}
    \label{fig:metrics}
\end{figure}

The image-level evaluation metrics, including PSNR, SSIM and Dice score of SDN are shown in Fig.\ref{fig:metrics}. Although PSNR and F1 are not favorable, this does not contradict the present study, which deliberately shifts the super-resolution objective from pixel-level assessment to bone microstructural inference.

\end{document}